%% file: main.tex
\documentclass[sigconf,natbib=true]{acmart}

\setcopyright{none}
\acmConference[Preprint]{arXiv preprint}{2026}{}
\acmISBN{}
\acmDOI{}
\renewcommand\footnotetextcopyrightpermission[1]{}

\usepackage[ruled,linesnumbered]{algorithm2e}
\usepackage{booktabs}
\usepackage{xcolor}

\begin{document}

\title{Constraint-Anchored Attribution: \\
       Feasibility-Certified Counterfactuals and Bonferroni-PAC
       Sufficient Subsets for Neural CO Policies}

\author{Sohaib Afifi}
\email{sohaib.afifi@univ-artois.fr}
\affiliation{%
  \institution{Univ. Artois, UR 3926, Laboratoire de G\'enie
               Informatique et d'Automatique de l'Artois (LGI2A)}
  \city{B\'ethune}
  \postcode{F-62400}
  \country{France}
}

\begin{abstract}
We give an attribution method for neural combinatorial-optimisation
(CO) policies that (i) decomposes a decision by constraint
families via LP-relaxation duals, (ii) certifies counterfactuals
through a combinatorial feasibility model (implemented as a
CSP feasibility-decision model), and
(iii) bounds the size of a PAC-sufficient explanation with a
Bonferroni-corrected Hoeffding sufficient-subset test along a
greedy ordering. Across three CO problems and three seeds, our
LP-anchored $\Lambda$-attribution matches the CF-derived signal
at $96.5\%$ on CVRPTW ($n_{\mathrm{cert}}{=}344$) and $77.2\%$ on
OP ($n_{\mathrm{cert}}{=}281$) vs $75.0\%$ and $35.2\%$ for proxy
gradient (diffs $+0.215$ and $+0.420$; McNemar
$p \leq 10^{-14}$). In the rank-aligned FJSP regime, both
backends agree on every CSP-certified flip
($n_{\mathrm{cert}}{=}59$), confirming the no-gain prediction.
Bonferroni-PAC subsets average $5.0$ nodes per step ($M{=}70$,
$\varepsilon{=}\delta{=}0.2$, $k_{\max}{=}25$). Code:
\url{https://github.com/sohaibafifi/neuro-co-cax}.
\end{abstract}

\keywords{explainable AI, combinatorial optimisation,
constraint programming, counterfactual explanation, neural routing}

\maketitle

\section{Introduction}
Neural combinatorial-optimisation (CO) policies now compete
with classical solvers on routing
\cite{kool2019attention,kwon2020pomo,berto2024rl4co,jari2025guni}
and scheduling \cite{cappart2023combinatorial}. As they move from
benchmarks to deployment, the question \emph{why did the policy
choose this action at this step?} becomes operationally
important: a dispatcher needs to know whether a route was
chosen because of a tight time window, a near-capacity vehicle,
or a geometric shortcut.

\paragraph{Why generic XAI fails on CO.}
Post-hoc feature-attribution methods such as
gradient~$\times$~input and gradient-axiomatic variants
(integrated gradients~\cite{sundararajan2017integrated},
DeepLIFT~\cite{shrikumar2017learning}) attribute each decision
to the input \emph{features} at the node level. A CO problem,
however, is defined by its \emph{constraint families}, each
typically spanning several feature tensors of different ranks.
As Section~\ref{sec:exp} shows on routing, a high-rank feature
tensor accumulates more gradient mass than a low-rank one even
when the underlying constraint family is not the binding one
at the LP optimum, so feature-level attribution mis-ranks the
constraint that actually drives the decision.

\paragraph{Why existing CO-specific explainers fall short.}
The closest prior work, RouteExplainer~\cite{kikuta2024routeexplainer},
trains a per-edge classifier on counterfactual labels for
Vehicle Routing Problem variants and narrates the labels with
an LLM. The counterfactuals it generates are unconstrained: on
$\approx 81\%$ of our routing instances (Section~\ref{sec:exp})
they push demand below zero or invert a time window, and the
explanation surface is
restricted to edges in a routing solution rather than the
underlying CO constraints. PAC
explanations via constraint programming~\cite{koriche2024model}
and abduction-based explanations~\cite{ignatiev2019abductive}
target generic classifiers, and formal CP-based explanations of
neural policies~\cite{selvey2023formal} target classical
planning domains, not the autoregressive CO decoder we study. Other recent threads in XAI-for-CO either focus on
surrogate models of static optimisation instances (e.g.,
SHAP-style attribution on a knapsack
ML-surrogate~\cite{garn2025transparency}) or remain pre-print;
none, to our knowledge, decomposes a neural CO policy's
decisions by constraint family with a CP-grounded feasibility
certificate.

\paragraph{Our contribution: Constraint-Anchored Attribution
(CAX).} Three primitives that connect neural explanations to the optimisation structure of the underlying CO problem:

\begin{enumerate}
\item \textbf{$\Lambda$-attribution}
      (Section~\ref{sec:method:lambda}): decompose a step's
      gradient by constraint family, scaled by the LP-relaxation
      Lagrangian multiplier $\lambda_k^*$. Yields
      \emph{constraint-level} attribution that no feature-level
      method can produce.
\item \textbf{Bonferroni-corrected PAC sufficient subset}
      (Section~\ref{sec:method:subset}): adapt
      Koriche et~al.~\cite{koriche2024model} to the CO decoder
      to give a family-wise Hoeffding-controlled top-$k$ size
      along the greedy attribution ordering, against which any
      attribution method can be scored.
\item \textbf{CSP-certified counterfactual}
      (Section~\ref{sec:method:cf}): lowest-$L_1$ perturbation
      found by the sampler that flips the argmax while satisfying
      the problem's CSP feasibility-decision model. The
      certificate distinguishes CAX from
      Wachter~\cite{wachter2017counterfactual} and
      Tsirtsis et~al.~\cite{tsirtsis2023counterfactual}.
\end{enumerate}

\noindent
The framework only requires a feasibility model; in our
experiments, all counterfactual certificates are implemented as
CSP feasibility-decision models.

\noindent
The counterfactual primitive in~(3) (henceforth abbreviated
CF) additionally produces a \emph{CF-derived adjudication
signal} that distinguishes three candidate backends for the
$\Lambda$ multipliers; on the benchmark of
Section~\ref{sec:exp} this signal is invariant under the norm
($L_1$/$L_2$/$L_\infty$) and per-family-dimension normalisation
of the perturbation mass, supporting its use as a reference
for ranking attribution backends.
The implementation runs on a single CPU and adds one post-hoc
feasibility-oracle call per decoding cell.\footnote{Code:
\url{https://github.com/sohaibafifi/neuro-co-cax} (MIT).}

\section{Method}
\label{sec:method}

\paragraph{Setup.}
Let $\mathcal{P}$ be a CO problem with $N$ \emph{nodes}
(customers, operations, \ldots) and constraint families
$\mathcal{C} = \{c_1, \ldots, c_K\}$. An instance is described
by feature tensors $x = (x^{(1)}, \ldots, x^{(F)})$ with
total dimension $d = \sum_i |x^{(i)}|$; each family $c_k$ is
parameterised by a subset $\mathcal{F}(c_k) \subseteq \{1,
\ldots, F\}$ of feature tensors. A learned policy $\pi_\theta$
unrolls over $T$ decoding steps, producing at step $t$ a
state $s_t$ (function of $x$ and the prefix of actions
$a_0, \ldots, a_{t-1}$) and a distribution over feasible
actions $a \in \mathcal{A}(s_t)$. We write
$a^\star(x) := \arg\max_a \pi_\theta(a \mid s_0(x))$
for the greedy first action induced by instance $x$. We assume
$\mathcal{P}$ admits an LP relaxation
$\widetilde{\mathcal{P}}(x)$ whose dual yields a per-row
multiplier vector; we aggregate to one scalar per family by
taking the mean absolute dual over the rows in that family,
giving $\lambda^*(x) = (\lambda_1^*, \ldots, \lambda_K^*)
\in \mathbb{R}_{\geq 0}^K$. Sensitivity to the aggregation
choice (mean vs sum vs max) is reported in
Section~\ref{sec:exp:ablation}.

\subsection{$\Lambda$-attribution}
\label{sec:method:lambda}
We attribute decision $a_t$ to constraint family $c_k$ as
\begin{equation}
  \Lambda_k(t) = \lambda_k^*(x) \cdot
    \sum_{i \in \mathcal{F}(c_k)} \sum_{j}
    \bigl| \nabla_{x^{(i)}_{j}} \log \pi_\theta(a_t \mid s_t)
           \cdot x^{(i)}_{j} \bigr|,
  \label{eq:lambda}
\end{equation}
where $j$ ranges over the scalar entries of feature tensor
$x^{(i)}$.
$\Lambda_k(t)$ is non-negative; the family with the largest
$\Lambda_k(t)$ is the one whose binding pressure most explains
the policy's preference for $a_t$. Three backends estimate $\lambda^*$:
\textsc{lp} (dual values of the LP relaxation
$\widetilde{\mathcal{P}}$),
\textsc{subgrad} (Lagrangian subgradient ascent on a relaxed
subproblem~\cite{fisher1981lagrangian}), and
\textsc{proxy} ($\lambda_k = 1$, which recovers a
constraint-partitioned gradient~$\times$~feature). The cost
of $\Lambda$ is one backward pass per decoding step plus $K$
per-family masked aggregations of $\nabla \log \pi \cdot x$;
our v0.2 reference implementation runs $K$ separate backwards
for clarity, which is the cost reported below and is trivially
reducible. The LP solve on $\widetilde{\mathcal{P}}$ is
shared across all decoding steps of an instance and amortises
over $T$.

\subsection{Bonferroni-PAC sufficient subset}
\label{sec:method:subset}
For PAC parameters $(\varepsilon, \delta) \in (0, 1)^2$, we
want the smallest node subset $S$ preserving the policy's
argmax under a Gaussian neighbourhood of $x$:
\begin{equation}
\begin{aligned}
  \min_S\;|S| \;\text{s.t.}\;
  \Pr_{x' \sim \mathcal{N}(x,\sigma^2 I)}\!\!
  \bigl[ a^\star(x'_{|S}) = a^\star(x') \bigr]
  \geq 1 - \varepsilon,
\end{aligned}
\label{eq:subset}
\end{equation}
where $x'_{|S}$ masks feature entries at nodes $\notin S$ to a
baseline value. A greedy variant walks an attribution ordering
and accepts the smallest $k$ at which the empirical
preserved-argmax rate exceeds $1 - \varepsilon$. The procedure
inspects up to $k_{\max}$ candidate subsets adaptively, so we
apply the Bonferroni union bound to obtain a family-wise
$(1-\delta)$-PAC guarantee along the greedy ordering: replacing
$\delta$ by $\delta/k_{\max}$ in the Hoeffding bound gives the
sample size
\begin{equation}
  M_{\mathrm{bonf}}
  \geq \left\lceil
  \frac{\log\!\bigl(2 k_{\max} / \delta\bigr)}{2\varepsilon^2}
  \right\rceil
  \label{eq:pac-m}
\end{equation}
per per-$k$ test, which we use throughout. The output is a
PAC-sufficient subset along the greedy ordering, not a
PAC-optimal subset in the exact-search sense; closing the
exact-search gap requires a CP encoding of the $M$-shot
preserve-rate test (Section~\ref{sec:discussion}) and is open.

\subsection{CSP-certified counterfactual}
\label{sec:method:cf}
Wachter-style counterfactuals~\cite{wachter2017counterfactual}
minimise a perturbation norm subject to an argmax flip, but do
not guarantee that the perturbed object remains a valid CO
instance. We add a combinatorial feasibility certificate: a
candidate perturbation is accepted only if the perturbed instance
satisfies the problem's CSP feasibility-decision model.
\begin{equation}
\begin{aligned}
  \min_{\zeta \in \mathbb{R}^d}\;& \lVert \zeta \rVert_1 \\
  \text{s.t.}\;& \zeta \in [-\rho, \rho]^d, \;
                \mathrm{feasible}_{\mathcal{P}}(x + \zeta), \\
              & a^\star(x + \zeta) \neq a^\star(x),
\end{aligned}
\label{eq:cf}
\end{equation}
with $\rho > 0$ a per-feature $L_\infty$ budget.
Equation~\eqref{eq:cf} defines the ideal counterfactual objective;
our implementation approximates it by sample-and-verify and returns
the lowest-$L_1$ CSP-certified perturbation found within the shot
budget, not a proof of global $L_1$ optimality.
The search is a two-stage sample-and-verify procedure.
The inner loop first applies cheap arithmetic checks
(problem-specific field-value bounds, $\sim 0.1$\,ms) and keeps
the lowest-$L_1$ flipping candidate encountered so far.
A post-hoc CSP feasibility-decision query (we use the CP-SAT
engine~\cite{perron2023cpsat}) is then issued on that candidate.
If the CSP model is satisfiable within the time limit, the
counterfactual is certified; otherwise the cell is rejected.
Timeouts or failures to certify are not interpreted as proofs
of infeasibility -- the certified subset is conservative.
Algorithm~\ref{alg:cf} gives the outer loop.

\begin{algorithm}[t]
\KwIn{policy $\pi_\theta$, env, instance $x$, feature keys
      $\mathcal{K}$, per-key $L_\infty$ bound $\rho$, per-key
      std $\sigma$, $M$ shots}
\KwOut{$\zeta^*_t$ CSP-certified counterfactual per step,
       or $\emptyset$}
\For{$t \in \{0, \ldots, T-1\}$}{
  $a_t \leftarrow \arg\max_a \pi_\theta(a \mid s_t)$;
  $\mathrm{best}\!\leftarrow\!\emptyset$\;
  \For{$m \in \{1, \ldots, M\}$}{
    $k \leftarrow m \bmod |\mathcal{K}|$
        \tcp*{round-robin one key at a time}
    $\zeta_k \sim \mathcal{N}(0, \sigma_k^2 I)
        \cap [-\rho_k, \rho_k]^{d_k}$;
    other keys: $\zeta = 0$\;
    \If{$\mathrm{feasible}^{\mathrm{arith}}_{\mathcal{P}}(x + \zeta)$
        \textbf{and} $\arg\max_a \pi_\theta(a \mid s_t(x + \zeta))
                    \neq a_t$
        \textbf{and} $\lVert\zeta\rVert_1
                       < \lVert\mathrm{best}\rVert_1$}{
      $\mathrm{best} \leftarrow \zeta$\;
    }
  }
  \tcp*[h]{post-hoc CSP certification on the cell winner}\;
  \lIf{$\mathrm{best} \neq \emptyset$ \textbf{and not}
       $\mathrm{CSP\text{-}feasible}_{\mathcal{P}}(x + \mathrm{best})$}{
       $\mathrm{best} \leftarrow \emptyset$
    }
  $\zeta^*_t \leftarrow \mathrm{best}$\;
}
\caption{Sample-and-verify counterfactual: arithmetic filter
inside the loop, CSP feasibility-decision certification post-hoc
on the per-cell winner.}
\label{alg:cf}
\end{algorithm}

\paragraph{Adjudication.}
The CF in eq.~\eqref{eq:cf} yields a CF-derived
\emph{adjudication signal}: on a CSP-certified flipping
perturbation, the family carrying the largest $|\zeta|$ mass
is the one to which the policy is most locally sensitive at
that step. Section~\ref{sec:exp} reports that this signal is
invariant under $L_1/L_2/L_\infty$ proximity and under
per-family-dimension normalisation across all $54$ sweep
configurations we tried; we therefore use it to score the
three $\Lambda$ backends head-to-head.

\section{Experiments}
\label{sec:exp}

\paragraph{Problems.}
We evaluate on three CO problems with contrasting constraint
profiles. \emph{Routing}: the \emph{Capacitated Vehicle
Routing Problem with Time Windows} (CVRPTW) is
constraint-rich: a fleet of identical vehicles must serve
customers with demands and time windows without exceeding
capacity, minimising total travel; families are
\textsc{capacity}, \textsc{time-window}, \textsc{spatial}.
The \emph{Orienteering Problem} (OP) is a prize-collecting
single-route variant: visit a subset of customers under a
travel-budget cap, maximising collected prize; families are
\textsc{prize}, \textsc{budget}, \textsc{spatial}.
\emph{Scheduling}: the \emph{Flexible Job-Shop Scheduling
Problem} (FJSP) extends JSSP with machine-flexibility: each
operation has a set of eligible machines and a per-machine
processing time; the policy must both order operations and
assign them to a machine while minimising makespan. Families
are intra-job \textsc{precedence} (parameterised by
processing times) and machine \textsc{eligibility}
(parameterised by the number-of-eligible-machines vector).

\paragraph{Why these three.}
The trio is chosen so each problem falsifies a different CAX
prediction. CVRPTW and OP expose multiple constraint families
parameterised by feature tensors whose ranks are
\emph{misaligned} with the LP-binding family: the highest-rank
tensor (e.g. CVRPTW's $N{\times}2$ time-window pairs) is not
the binding one, so proxy attribution mis-ranks. They are our
\emph{adjudication substrates} on which the CSP-certified
counterfactual adjudication signal and LP-vs-proxy agreement are
computed
(Sections~\ref{sec:exp:lambda},~\ref{sec:exp:cf},~\ref{sec:exp:pac});
OP additionally cross-family-replicates the CVRPTW result on a
prize-collection variant. FJSP, in contrast, is multi-family
with its high-mass tensor (\textsc{eligibility}'s
eligible-machine count vector) \emph{aligned} with the
LP-binding family. CAX predicts that proxy and LP backends
\emph{agree} on top-1 family in this regime, so FJSP is our
\emph{theory-confirming ablation}
(Section~\ref{sec:exp:ablation}). Any \textsc{lp}-vs-proxy
discordance on FJSP would falsify CAX's account of when LP
duals add information.

\paragraph{Coverage and CSP certification.}
The counterfactual search first applies arithmetic feasibility
inside the sampling loop, then applies a CSP feasibility-decision
query post-hoc to the best flipping candidate in each cell.
Arithmetic filtering produces $344/384$ flips on CVRPTW,
$315/384$ on OP, and $59/384$ on FJSP (3 seeds, $B{=}16$,
$T{=}8$, $M{=}128$). Per-cell CSP certification accepts $344$,
$281$, and $59$ respectively; OP rejections come from the
generator's prize-positivity envelope. All headline attribution
numbers in Section~\ref{sec:exp:lambda} are computed on this
CSP-certified subset. A matched unconstrained counterfactual
baseline using the same Gaussian sampler but no feasibility
filter passes the arithmetic check on only $19.3\%$ of
candidates.

\paragraph{Robustness of the CF-derived signal.}
The CF top-family choice is invariant across norms
($L_1$/$L_2$/$L_\infty$) and per-family dimension
normalisation (which kills the ``family with most dimensions
wins'' artefact). Sweeping
$3$ seeds $\times$ $3$ norms $\times$ $2$ normalisations
$\times$ $3$ problems = $54$ configurations: top-$1$ CF
family is stable across all $54$ configurations tested (CVRPTW:
\textsc{spatial}, OP: \textsc{prize}, FJSP:
\textsc{eligibility}).

\paragraph{Setup.}
For CVRPTW and OP we use the rl4co~\cite{berto2024rl4co}
attention-based encoder-decoder~\cite{kool2019attention}
(CVRPTW: $N{=}50$ customers, $50$-start multistart-greedy
decoding; OP: $N{=}20$); for FJSP, rl4co's matrix-attention
backbone at $10$ jobs $\times$ $5$ machines. We train three
seeds per problem. Realised optimality gap (mean$\pm$std across seeds) against
problem-specific classical optimisation baselines is
$10.65{\pm}0.49\%$ on CVRPTW,
$10.21{\pm}0.26\%$ on OP, and $7.09{\pm}0.08\%$ on FJSP;
all three policies are within the $15\%$ band the rl4co
reference implementations report for analogous schedules.
Realised optimality gaps are computed against problem-specific
classical optimisation baselines; these baselines are separate
from the CSP feasibility-decision models used for counterfactual
certification. CAX runs on a single CPU; the heaviest primitive
(the CSP-certified counterfactual procedure with $128$ shots) costs
$\sim 30$\,s per (run dir, $B{=}16$, $T{=}12$).

\subsection{$\Lambda$-attribution vs CSP-certified counterfactual adjudication signal}
\label{sec:exp:lambda}

Across all three seeds with $B{=}16$, $T{=}8$, $M{=}128$ CF
shots and post-hoc CSP certification, the search yields
$n_{\mathrm{cert}}{=}344$ CSP-certified flipping perturbations
on CVRPTW ($100\%$), $281$ on OP ($89.2\%$), and $59$ on FJSP
($100\%$). All headline numbers below use only the certified
subset. Every backend (including the feature-attribution
\textsc{proxy}) reduces to a per-cell top-$1$ constraint family
by partitioning $|\nabla \log\pi \cdot x|$ over the family's
feature tensors and applying the backend's $\lambda_k$
weighting, so the agreement metric below is a fair comparison
of family-level rankings. Mean agreement with the CF-derived
signal (mean$\pm$std across seeds):

\begin{center}
\begin{tabular}{lrrr}
\toprule
backend & CVRPTW & OP & FJSP \\
\midrule
\textsc{proxy}   & $0.75{\pm}0.02$         & $0.35{\pm}0.06$         & $1.00{\pm}0.00$ \\
\textsc{lp}      & $\mathbf{0.97{\pm}0.01}$ & $\mathbf{0.77{\pm}0.04}$ & $1.00{\pm}0.00$ \\
\textsc{subgrad} & $0.00$ & $0.00$ & $0.00$ \\
\bottomrule
\end{tabular}
\end{center}

\begin{figure}[t]
  \centering
  \includegraphics[width=\linewidth]{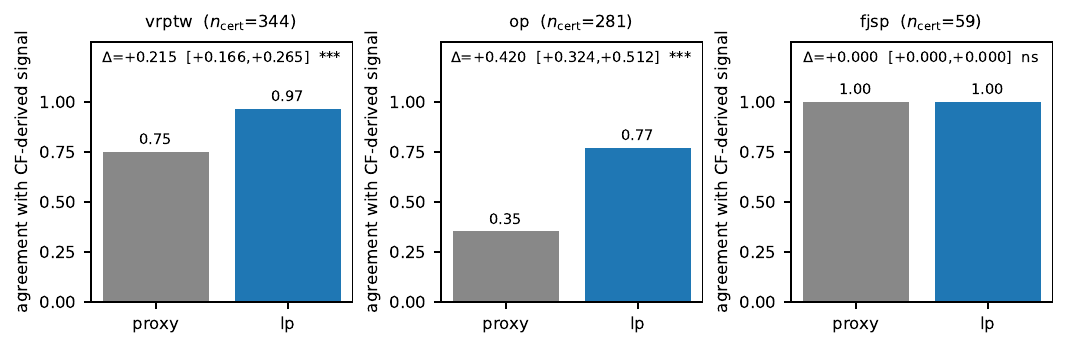}
  \caption{Mean agreement between $\Lambda$-attribution top-1
    family and the CF-derived signal on CSP-certified
    counterfactual cells, per backend, on CVRPTW, OP, and FJSP
    ($n_{\mathrm{cert}}{=}344, 281, 59$
    respectively). Annotations: paired bootstrap $95\%$ CI on
    the diff vs \textsc{proxy} and McNemar exact $p$-value
    (*** $< 10^{-3}$, ns for indistinguishable backends).}
  \label{fig:adjudication}
\end{figure}

\noindent
On the pooled paired vectors (per-cell match indicators on
certified flips across seeds), LP $\Lambda$ beats proxy by
$+0.215$ on CVRPTW (bootstrap $95\%$ CI $[+0.166, +0.265]$,
$10{,}000$ resamples; McNemar $p = 4.5\times 10^{-17}$,
$b_{01}{=}7$, $b_{10}{=}81$) and by $+0.420$ on OP
($[+0.324, +0.512]$; $p = 9.3\times 10^{-15}$,
$b_{01}{=}60$, $b_{10}{=}178$). On FJSP, the rank-alignment
regime, both backends are exactly equal on the $59$
certified cells ($b_{01}{=}b_{10}{=}0$; $p = 1$) and both reach
$1.00$ agreement. CAX's prediction is supported empirically:
LP duals rescue attribution when feature ranks and the
binding family disagree, and add no information when they
align. Figure~\ref{fig:adjudication} shows the per-problem
breakdown. \textsc{subgrad} recovers the LP-binding family on
the relaxed subproblem (budget on OP, capacity on CVRPTW) but
puts its weight on a low-rank feature tensor; the aggregate
$\Lambda$ ranking flips, so we keep it as a negative control.

\subsection{Counterfactual feasibility rate}
\label{sec:exp:cf}
Arithmetic filtering runs inside the sampling loop on every
shot; the combinatorial oracle (Section~\ref{sec:method:cf})
is invoked once per cell on the per-cell winner. Headline
Section~\ref{sec:exp:lambda} numbers are on the
combinatorially-certified subset only; the arithmetic stage
alone is reported here for comparison. Realised arithmetic
pass-rate on the per-cell winners is $100\%$ on all three
problems by construction (CVRPTW, OP, FJSP). A matched
Wachter baseline (same Gaussian sampler, no feasibility
filter) on the same seeds passes the arithmetic check on
only $19.3\%$ of candidates: the joint perturbation pushes
demand below zero or inverts a time-window pair on the
remaining $\approx 81\%$.

\subsection{Bonferroni-corrected PAC sufficient subset on CVRPTW}
\label{sec:exp:pac}
We run eq.~\eqref{eq:subset} on the CVRPTW checkpoint
(seed 0) with $(\varepsilon, \delta) = (0.2, 0.2)$,
$\sigma = 0.05$, $k_{\max} = 25$. The Bonferroni-tight
sample size for $k_{\max}$ adaptive queries
($M_{\mathrm{bonf}} = \lceil \log(2 k_{\max}/\delta) /
(2\varepsilon^2) \rceil = 70$) gives a family-wise
$(1{-}\delta)$-PAC guarantee along the greedy ordering;
the per-test budget ($M = 29$) is the un-corrected analogue.
The greedy procedure succeeds on $40$ of $64$ cells with
mean $|S^*| = \mathbf{4.97}$ nodes (median $2$, max $24$).
The remaining $24$ cells stay below preserved-rate $0.20$
at every $k \leq k_{\max}$; their top-$1$ vs top-$2$ logit
margin median is $0.36$ (vs $0.83$ on succeeded cells), so
failures concentrate on near-tied decision states whose
argmax is noise-sensitive. Bonferroni and per-test bounds
return identical success counts and identical $|S^*|$ because
the larger sample size tightens the concentration without
moving point estimates.

\subsection{LP-vs-proxy backend ablation across problems}
\label{sec:exp:ablation}
On CVRPTW and OP, the two backends disagree on top-family
ranking and the CF-derived adjudication signal of
Section~\ref{sec:exp:lambda} sides with \textsc{lp}: the
$b_{10}{:}b_{01}$ ratios ($81{:}7$ on CVRPTW,
$201{:}67$ on OP) decompose the McNemar effect as
``LP-rescues-proxy mistakes'' rather than the symmetric case.
On FJSP both backends pick \textsc{eligibility} on every
cell: the LP shadow price for the assignment row dominates
the precedence-row shadow price by an order of magnitude,
and the eligible-machine count tensor also dominates the
proxy gradient mass. CAX predicts this regime should yield
zero LP advantage; the $b_{01}{=}b_{10}{=}0$ result on $59$
CSP-certified flips is a sharp empirical confirmation.

\paragraph{Why mean aggregation.}
Pair-wise top-$1$ family agreement under mean/sum/max
aggregation is $1.00$ on FJSP and $\geq 0.875$ on CVRPTW. On
OP, mean and sum disagree completely ($0.00$): the spatial
family has $O(N^2)$ LP rows (degree + MTZ) versus $O(1)$ for
the prize family, and sum scales linearly with row count.
Mean aggregation avoids row-count bias and is the aggregation that recovers the CF-derived signal's prize-binding pick on OP. We therefore adopt mean as the canonical aggregator and
report sum/max as ablations.

\section{Discussion and conclusion}
\label{sec:discussion}

\paragraph{Limitations.}
Bonferroni-PAC walks the greedy ordering, not the full subset
lattice. The CSP certificate covers only accepted CFs:
timeout = rejection, not proof of infeasibility, so the
certified subset is conservative. The Beasley subgradient
backend degenerates on heuristically-feasible relaxed
subproblems; kept as negative control. Single-family
substrates (basic JSSP) admit no CF flips under the rl4co
generator; FJSP is the multi-family scheduling analogue.

\paragraph{Perspectives.}
(a) branch-and-bound + attribution-ordered upper bounds for
exact PAC-optimal $|S^*|$; (b) broader CO suite (TSP, MTSP,
knapsack, set cover, RCPSP, graph colouring) to map
rank-mismatch vs rank-alignment regimes; (c) prefix-constrained
regret oracle as a continuous CF signal on rank-tight problems.

\paragraph{Conclusion.}
CAX decomposes a neural CO policy's decisions by binding
constraint families, certifies counterfactuals via a per-cell
CSP feasibility-decision query, and bounds PAC-sufficient subset
size along a greedy ordering with a Bonferroni-corrected
Hoeffding test. On CSP-certified flips,
LP-anchored $\Lambda$ matches the CF-derived signal at $97\%$
vs $75\%$ for proxy on CVRPTW ($+0.215$,
$p = 4.5\times 10^{-17}$, $n_{\mathrm{cert}}{=}344$) and at
$77\%$ vs $35\%$ on OP ($+0.420$,
$p = 9.3\times 10^{-15}$, $n_{\mathrm{cert}}{=}281$); both
backends agree perfectly on FJSP ($n_{\mathrm{cert}}{=}59$),
the rank-aligned regime. LP duals rescue attribution when
feature ranks and LP-binding disagree, and add no information
when they align.

\bibliographystyle{ACM-Reference-Format}
\input{main.bbl}

\end{document}